\pdfoutput=1  

\documentclass[letterpaper, 10 pt, conference]{ieeeconf}

\IEEEoverridecommandlockouts
\usepackage{amsmath}
\usepackage{amssymb}
\usepackage{graphicx}
\usepackage{booktabs}
\usepackage{array}
\usepackage{url}
\usepackage{tikz}
\usetikzlibrary{arrows.meta, positioning, fit, backgrounds, calc}

\title{\LARGE \bf
Toward Certified Functional Safety for Industrial Humanoid Robots:\\
The Fail-Passive Gap and a Feasibility Study
}

\author{Caiwu Ding, Tao Cui, Lingyun Wang, and Chengtao Wen
\thanks{The authors are with Siemens Foundational Technologies, Siemens Corporation,
        Princeton, NJ, USA
        {\tt\small \{caiwu.ding, tao.cui, lingyun.wang, chengtao.wen\}@siemens.com}}%
}

\begin{document}

\maketitle
\thispagestyle{empty}
\pagestyle{empty}


\renewcommand{\topfraction}{0.92}
\renewcommand{\bottomfraction}{0.85}
\renewcommand{\textfraction}{0.06}
\renewcommand{\floatpagefraction}{0.72}
\setcounter{topnumber}{3}
\setcounter{bottomnumber}{2}
\setcounter{totalnumber}{4}

\begin{figure*}[!t]
\centering
\resizebox{0.92\textwidth}{!}{%
\begin{tikzpicture}[
    font=\normalsize,
    box/.style={draw, rounded corners=2pt, align=center, inner sep=4pt,
                minimum height=9mm, text width=24mm, fill=white},
    sbox/.style={draw, rounded corners=2pt, align=center, inner sep=4pt,
                 minimum height=9mm, text width=27mm},
    flow/.style={-{Latex[length=2.4mm]}, very thick},
    hazard/.style={-{Latex[length=2.4mm]}, very thick, red!75!black},
    panel/.style={draw, rounded corners=4pt, inner sep=9pt},
    ptitle/.style={font=\bfseries, align=center}
]
\node[box] (l-cls) {industrial arms,\\cobots, AGVs};
\node[box, right=10mm of l-cls] (l-es) {E-stop /\\light curtain};
\node[box, right=10mm of l-es] (l-ct) {contactors\\(Stop Cat.~0)};
\node[box, right=10mm of l-ct] (l-pw) {power\\removed};
\node[sbox, right=10mm of l-pw, fill=green!14] (l-safe) {\textbf{de-energized}\\\textbf{safe state} ---\\\textbf{CERTIFIABLE}};
\draw[flow] (l-cls) -- (l-es);
\draw[flow] (l-es) -- (l-ct);
\draw[flow] (l-ct) -- (l-pw);
\draw[flow] (l-pw) -- (l-safe);
\node[below=3mm of l-ct, font=\small\itshape, text=green!45!black, align=center]
     (l-std) {certified PL\,e / SIL\,3\\ISO 13849-1, IEC 62061,\\ISO 10218, ISO 3691-4};
\begin{scope}[on background layer]
\node[panel, draw=green!55!black, fit=(l-cls)(l-es)(l-ct)(l-pw)(l-safe)(l-std)] (Lpanel) {};
\end{scope}
\node[ptitle, anchor=west] at ([yshift=2.5mm]Lpanel.north west)
     {(a) Classic machinery --- \textit{fail-passive}: ``stop'' is unconditionally safe, and \textbf{certifiable} today};

\node[box, below=30mm of l-cls] (r-cls) {industrial\\humanoid (biped)};
\node[box, right=10mm of r-cls] (r-es) {E-stop /\\light curtain};
\node[box, right=10mm of r-es] (r-ps) {wireless\\PROFIsafe};
\node[box, right=10mm of r-ps] (r-pol) {robot control\\policy (SDA\\reception)};
\node[sbox, right=10mm of r-pol, fill=yellow!22] (r-safe) {\textbf{actively balanced}\\\textbf{standstill} ---\\\textbf{UNCERTIFIED}};
\draw[flow] (r-cls) -- (r-es);
\draw[flow] (r-es) -- (r-ps);
\draw[flow] (r-ps) -- (r-pol);
\draw[flow] (r-pol) -- (r-safe);
\node[sbox, below=8mm of r-pol, fill=red!10, draw=red!70!black, text width=32mm]
     (r-fall) {\textcolor{red!75!black}{\textbf{remove power $\Rightarrow$ fall = hazard}}};
\draw[hazard] (r-pol.south) -- (r-fall.north);
\node[below=3mm of r-es, font=\small\itshape, text=red!70!black, align=center]
     (r-std) {no certified\\reaction pathway};
\begin{scope}[on background layer]
\node[panel, draw=red!70!black, fit=(r-cls)(r-es)(r-ps)(r-pol)(r-safe)(r-fall)(r-std)] (Rpanel) {};
\end{scope}
\node[ptitle, anchor=west] at ([yshift=2.5mm]Rpanel.north west)
     {(b) Balancing humanoid --- \textit{active safe state}: ``stop'' must be \textit{controlled}, and is \textbf{not yet certifiable}};
\node[below=2mm of Rpanel.south, font=\bfseries\color{red!75!black}, align=center]
     {the \textbf{fail-passive gap}: classical de-energization is itself a hazard for a walking biped};
\end{tikzpicture}%
}
\caption{The \emph{fail-passive gap}. (a)~Classical machinery---fixed industrial
arms, collaborative robots, and AGVs---reaches the safe state by \emph{de-energizing};
a protective stop removes power and the machine coasts to a harmless standstill
(Stop Category~0), which is \emph{certifiable} today under ISO 13849-1 / IEC 62061
(and robot/vehicle standards ISO 10218, ISO 3691-4) to PL\,e / SIL\,3. (b)~A
dynamically balancing humanoid inverts this: removing power causes an uncontrolled
fall, so the safe state is an \emph{actively-controlled} balanced standstill that
depends on the robot's control policy. The mechanism classical safety relies on
becomes a hazard source, and no certified reaction pathway yet exists---the gap this
paper characterizes.}
\label{fig:teaser}
\end{figure*}

\begin{abstract}
Industrial humanoid robots are constrained less by locomotion or manipulation
capability than by the immaturity of functional safety certification for legged
platforms. The root difficulty is that \emph{the safe state of a legged robot is
an actively-controlled state, which violates the fail-passive assumption
underlying ISO~13849-1 / EN~60204-1}: removing power from a walking biped causes an
uncontrolled fall, so classical de-energization is itself a hazard. We term this
the \emph{fail-passive gap} and use a certified external safety chain (light
curtain, emergency stop, fail-safe input, fail-safe PLC, and wireless PROFIsafe) as
an instrument to locate it precisely: because the external chain is closed and
quantifiable with established methods (PFHD, DC, CCF, PL/SILCL), the residual
uncertifiable element is pinpointed to the robot-side reaction chain. Using a
Siemens fail-safe S7-1500 emergency-stop reference, we show its certifiable
Reaction subsystem is contactor-based power removal (Stop Category~0)---exactly the
element a balancing humanoid cannot have. We deliberately do not claim end-to-end
certified PL~e / SIL~3. We validate the approach on a Unitree G1 EDU pick-and-place
cell in a $3\,\mathrm{m}\times1.5\,\mathrm{m}$ semi-enclosed workspace, and
contribute a humanoid-specific analysis of the active safe state (fall-as-hazard,
single-support stop bounds, balancing-policy residual risk, ISO~13855 separation)
and a provenance-labeled timing budget. Hosting an industrial
software-defined automation (SDA) controller \emph{on the robot}, co-located with
the balancing policy, moves robot-side PROFINET/PROFIsafe reception onto a
standardized IEC~61131-3 interface; because the G1's onboard compute is not
safety-rated hardware, this endpoint is \emph{not} a certified safety runtime,
which reinforces rather than resolves the fail-passive gap and localizes it to the
SDA-to-balancing-policy interface.
\end{abstract}

\section{Introduction}

The safe state of a legged robot is an actively-controlled
state~\cite{grandia2021multi}, which violates
the fail-passive assumption underlying ISO~13849-1 / EN~60204-1, where the safe
state is the de-energized state. In classical machinery safety a protective stop
removes power and the machine coasts or brakes to a harmless standstill; a Stop
Category~0 \emph{is} power removal. For a dynamically balancing humanoid this
assumption inverts: removing power from a walking biped produces an uncontrolled
fall, which is itself a hazard. The safe state is instead a controlled, actively
balanced standstill---an \emph{active safe state}---that depends on the robot's
own real-time control policy remaining functional. The mechanism that classical
safety relies on to guarantee safety therefore becomes a hazard source. We call
the resulting certification gap the \emph{fail-passive gap} (Fig.~\ref{fig:teaser}),
and characterizing it precisely is the central contribution of this paper.

Humanoid robots are increasingly proposed for industrial tasks designed around
the human body, and platforms such as the Unitree G1 EDU have made dynamic bipedal
locomotion and dexterous manipulation broadly accessible. Yet the barrier to
deployment near people is not capability but functional safety. For fixed-base
manipulators and wheeled mobile platforms this is supported by a mature standards
ecosystem (ISO~13849-1/-2 and IEC~62061 for safety-related control systems,
EN~ISO~13850 for emergency stop, EN~60204-1 for stop categories, ISO~10218 and
ISO/TS~15066 for collaborative operation). None of these resolves the fail-passive
gap for a balancing biped.

\textbf{Approach: use a certified external chain as an instrument to locate the
gap.} Rather than proposing new safety hardware, we build a conventional,
PL~e / SIL~3-capable external safety supervision chain (light curtain,
emergency stop, fail-safe input, fail-safe PLC, wireless PROFIsafe) around a
humanoid work cell, and assess it with established methods. Because this external
chain is closed and quantifiable using existing standards, it acts as a measuring
instrument: everything up to the point where the safe stop command leaves the PLC
is accounted for, so whatever remains uncertifiable can be pinpointed to exactly
one place---the robot-side reaction chain. The external chain is thus both a
practical, certifiable safety architecture and the instrument that makes the
humanoid gap precise.

We address two coupled questions: (i) how to build a practical, deployable safety
supervision system for a humanoid in a semi-enclosed industrial workspace using
certified components; and (ii) how to evaluate that safety function in a
certification-oriented way with established machinery standards, given that no
dedicated industrial humanoid safety standard yet exists. A key enabler is
\emph{where} the industrial safety endpoint lives: rather than terminating the
PROFIsafe channel in a fixed control cabinet, we host an industrial
software-defined automation (SDA) controller \emph{on the robot's own onboard
compute}, co-located
with the balancing policy. The novelty is this \emph{on-robot placement}: hosting
the runtime directly on the robot puts a standardized IEC~61131-3
safety-reception layer against the proprietary balancing controller, making the
legged reaction chain approachable by industrial functional-safety methods.
Consistent with the current absence of a humanoid-specific standard, we scope the
work as a certification-oriented feasibility study and gap analysis: we build and
evaluate the external chain to established methods, and precisely characterize---
rather than prematurely certify---the residual robot-side reaction chain. Within
this scope, primary risk reduction is the light-curtain protective stop (with the
emergency stop treated as a supplementary safety function per EN~ISO~13850 /
EN~60204-1).

\textbf{Contributions.}
\begin{itemize}
\setlength{\itemsep}{0pt}\setlength{\parskip}{0pt}
\item[\textbf{C1}] A standards-aware external functional safety supervision
architecture for an industrial humanoid cell, built entirely from certified
components and structured as a Detection--Evaluation--Reaction (D--E--R) chain
over wireless PROFIsafe.
\item[\textbf{C2}] Use of the certified external chain as an \emph{instrument}
that pinpoints the fail-passive gap to the robot-side reaction chain, supported
by a direct comparison with a Siemens fail-safe S7-1500 emergency-stop reference
whose Reaction subsystem is certifiable precisely because it is contactor-based
Stop Category~0.
\item[\textbf{C3}] A humanoid-specific safety analysis of the active safe state:
the fall-as-hazard trade-off, mid-step (single-support) stop demands, residual
risk of balancing-policy failure, and an ISO~13855 separation-distance treatment
tied to the light-curtain placement.
\item[\textbf{C4}] A populated system-level timing/latency budget distinguishing
specified, configured, and measured terms, yielding a worst-case reaction time
and reaction distance.
\item[\textbf{C5}] A transparent, reproducible experimental protocol with metrics
explicitly labeled by provenance (specified, configured, measured) and
recommended instrumentation.
\item[\textbf{C6}] A novel IT/OT-convergent robot-side architecture that relocates
the industrial safety endpoint \emph{onto the humanoid itself}: a software-defined
automation (SDA) controller (a soft PLC) is hosted on the robot's onboard compute,
co-located with
the locomotion/balancing policy, so that a standardized IEC~61131-3 reception layer
sits directly against the proprietary balancing controller---rather than bridging
back to a fixed control cabinet. This on-robot placement is what makes the legged
reaction chain reachable by industrial functional-safety methods. We further
delineate its boundary precisely: because the G1's onboard compute is not
safety-rated hardware, this endpoint standardizes the reception interface but is
\emph{not} a certified PROFIsafe F-host and carries no SIL/PL claim, which localizes
the fail-passive gap to the well-defined SDA-to-balancing-policy interface.
\end{itemize}

\section{Related Work}

\textbf{Machinery functional safety.} ISO~13849-1 provides the performance-level
(PL) framework (category, MTTF$_\mathrm{D}$, DC, CCF, PFHD)\footnote{For readers
outside functional safety: \emph{Category} (B,\,1--4) is the safety-function
architecture/redundancy class; \emph{MTTF$_\mathrm{D}$} the mean time to a
dangerous channel failure; \emph{DC} (diagnostic coverage) the fraction of
dangerous failures detected; \emph{CCF} the resistance to common-cause failures;
and \emph{PFHD} the probability of a dangerous failure per hour---together
yielding the discrete \emph{PL} (Performance Level a--e; PL~e highest). Our
external Detection--Evaluation--Reaction chain establishes all of these
(Category~4, DC~$\geq99\%$, CCF~$\geq65$ points $\Rightarrow$ PL~e), whereas the
humanoid reaction chain establishes none---no PFHD, hence no PL---which is the
fail-passive gap.}; IEC~62061 provides a
parallel SIL/SILCL framework~\cite{iso13849,iec62061}; ISO~13849-2 defines
validation; EN~60204-1 defines stop categories (0, 1, 2); and EN~ISO~13850
specifies the emergency-stop function as complementary rather than primary risk
reduction~\cite{estop}. These are mature and widely applied to guards, light
curtains, safety PLCs, and drive-based safe stopping.

\textbf{Safety-rated sensing and safe communication.} Type~4 electro-sensitive
protective equipment (ESPE) such as light curtains is standardized under
IEC~61496~\cite{iec61496}, and PROFIsafe (IEC~61784-3) provides a ``black-channel''
safety layer for SIL~3 signaling over standard (including wireless) networks via
sequence numbering, watchdog timing, and CRC~\cite{profisafe}. Wireless PROFIsafe
over industrial WLAN (e.g., SCALANCE~W) is established for mobile applications,
subject to worst-case latency and watchdog configuration.

\textbf{Software-defined automation (SDA).} IEC~61131-3 logic increasingly
runs on software PLCs hosted on standard x86/x64 industrial PCs and edge devices,
often containerized on Linux, reflecting the broader IT/OT-convergence trend. Such
software-defined automation (SDA) controllers execute compiled IEC~61131-3
programs on Linux and communicate with field devices over standard protocols
(PROFINET, OPC~UA) and, in a safety configuration, PROFIsafe. To date such
runtimes are deployed on fixed industrial PCs and edge devices; hosting one
\emph{on a mobile
robot itself}, co-located with the locomotion controller, is comparatively
unexplored and directly relevant to legged platforms. In this placement the
industrial safety endpoint travels with the robot and interfaces directly with the
balancing policy, rather than bridging back to an external cabinet controller. This
on-robot placement is central to our robot-side architecture and is what lets
industrial functional-safety methods reach the legged reaction chain.

\textbf{Collaborative and mobile robot safety.} ISO~10218-1/-2 (revised 2025) and
ISO/TS~15066 define collaborative operation modes---safety-rated monitored stop,
hand-guiding, speed and separation monitoring (SSM), and power and force limiting
(PFL); ISO~13855 specifies how protective-equipment placement relates to approach
speed and stopping performance, and ISO~3691-4 addresses driverless industrial
trucks / AMRs~\cite{robotstd,iso13855}. A substantial body of research operationalizes these modes: SSM
using safety-rated vision and depth sensing to modulate robot speed with human
proximity and dynamic separation-distance computation~\cite{marvel2013ssm,byner2019dynamic},
and PFL biomechanical-limit studies underlying the ISO/TS~15066
thresholds~\cite{haddadin2017collisions}. These provide partial conceptual
analogs but share a structural assumption: the platform's safe state is
fail-passive---reached by stopping or de-energizing---and its stopping dynamics
are well characterized. That assumption is exactly what a balancing biped
violates, so SSM/PFL results transfer only up to the point of \emph{demanding} a
stop, not to \emph{executing} one safely on a legged platform.

\textbf{Humanoid and legged-robot safety.} Research on legged robots emphasizes
push recovery and balance control, capture-point / capturability-based
stepping~\cite{pratt2006capture,koolen2012capturability}, fall detection and
prediction, damage-mitigating or compliant
falling~\cite{fujiwara2002ukemi,ha2015fall}, and compliant (series-elastic /
variable-impedance) actuation~\cite{pratt1995sea,ham2009compliant}. This
literature targets keeping the robot upright or minimizing fall damage, not
certified functional safety of a commanded protective stop. A recent
learning-based approach reframes the humanoid stop as reaching a minimum-risk
condition via a fallback controller, training a neural monitor that predicts
real-time safe-stoppability and intervenes before safe stopping becomes
infeasible~\cite{sun2026learning}; this targets \emph{executing} the stop, but
yields a probabilistic confidence rather than a certifiable guarantee. Recent
industrial-humanoid pilots (e.g., Agility Digit, Figure, Apptronik
Apollo)~\cite{saeedvand2019survey} intensify the need, yet no dedicated,
harmonized industrial humanoid safety standard governs the reaction chain of a
balancing biped. We contribute by transferring the machinery-safety toolkit as far
as it credibly goes and marking the boundary---the fail-passive gap---where it
stops.

\section{System Architecture}

\subsection{Application Scenario}
\begin{figure}[t]
\centering
\includegraphics[width=1.0\columnwidth]{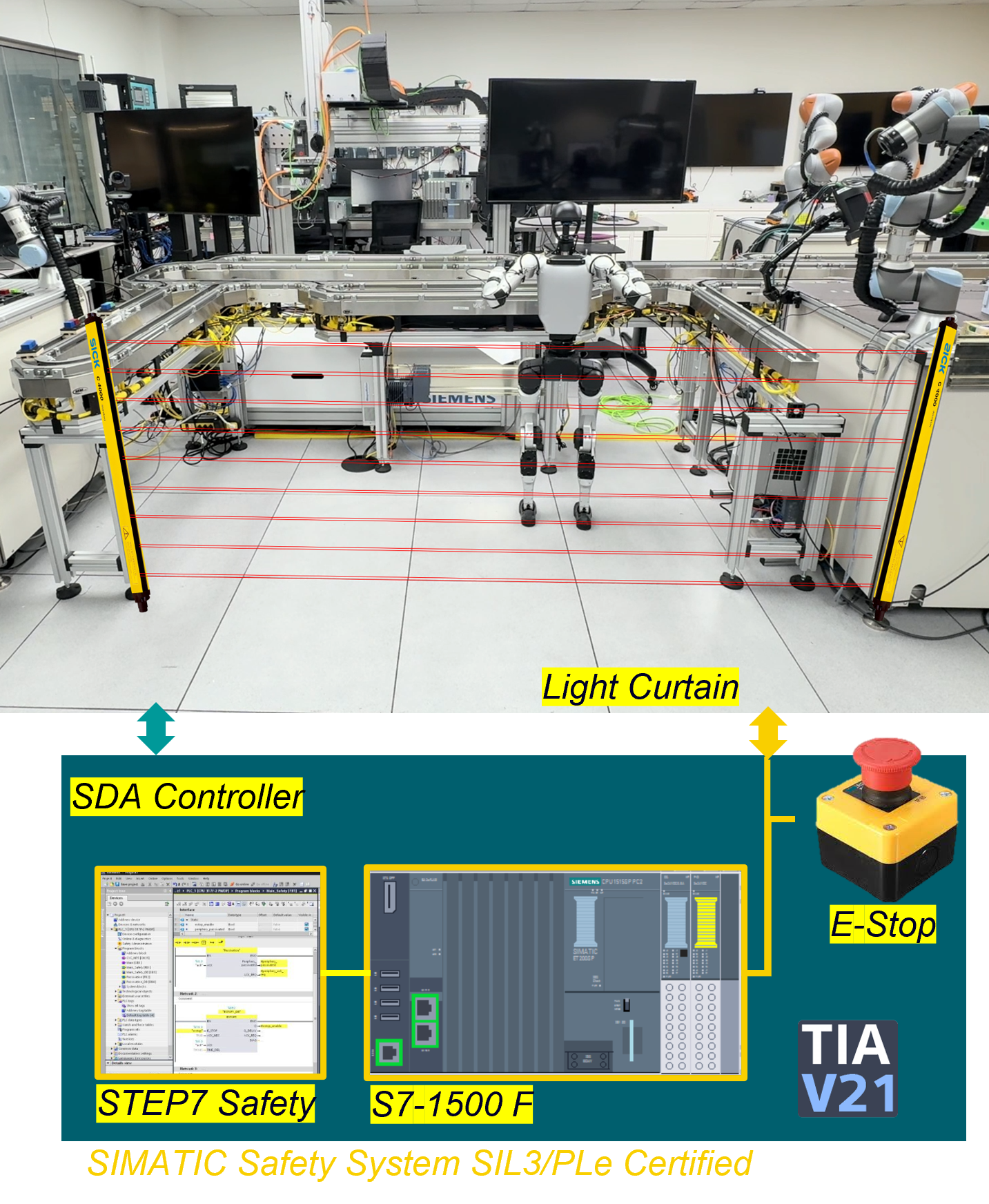}
\caption{The semi-enclosed conveyor cell and the external fail-safe (F-PLC)
supervision chain. The conveyor geometry forms a partial enclosure; the open
(south) side is monitored by the SICK deTec2 Core light-curtain pair, and the
Siemens fail-safe PLC (Detection\,+\,Evaluation) issues the safe stop demand
conveyed to the robot over wireless PROFIsafe.}
\label{fig:cell}
\end{figure}

A Unitree G1 EDU humanoid operates between two waypoints/workstations along a
conveyor in an approximately $3\,\mathrm{m}\times1.5\,\mathrm{m}$ semi-enclosed
workspace (Fig.~\ref{fig:cell}). Its task is a simple industrial pick-and-place:
picking up and dropping a $1\times1$~inch cube. Locomotion uses the robot's native
low-body walking control policy. The cell is semi-enclosed: the conveyor geometry
forms a partial physical enclosure on most of the perimeter, while one open side
(south) remains accessible to operators. That open side is monitored by a
light-curtain pair. Operators can start or stop the task flow; the workflow is
orchestrated automatically by the PLC.

\subsection{Hardware Overview}
The design uses \emph{no external safety relays or contactors}: the stop demand is
realized through the fail-safe PLC and conveyed to the robot as a safety telegram,
not by hard-wired power interruption. This is a deliberate architectural
consequence of the legged platform and is central to the certification discussion
(Sec.~\ref{sec:eval} and Sec.~\ref{sec:limits}). Table~\ref{tab:hw} lists the
components~\cite{vendordocs}.

\begin{table}[htbp]
\caption{System hardware inventory}
\label{tab:hw}
\begin{center}
\footnotesize
\renewcommand{\arraystretch}{1.0}
\begin{tabular}{|p{2.6cm}|p{5.0cm}|}
\hline
\textbf{Function} & \textbf{Component}\\
\hline
Safety PLC/controller & SIPLUS ET 200SP Open Controller, SIPLUS CPU 1515SP PC2 F\\
\hline
Safe input module & ET 200SP F-DI 8$\times$24VDC HF\\
\hline
Protective sensing & SICK deTec2 Core light curtain (1213210/1213211), 1 Tx + 1 Rx\\
\hline
Emergency stop & Siemens 3SU1851-0NB00-2AA2\\
\hline
Safety relays/contactors & none\\
\hline
Wireless safe comms & SCALANCE W (wireless PROFIsafe / black channel)\\
\hline
Robot-side runtime & Software-defined automation (SDA) controller (soft PLC) on G1 onboard (non-safety-rated) Linux compute; IEC 61131-3; robot-side PROFIsafe endpoint (not a certified F-host)\\
\hline
Engineering & TIA Portal; SDA engineering tooling\\
\hline
\end{tabular}
\end{center}
\end{table}

\subsection{Logical Architecture and Safety Zone}
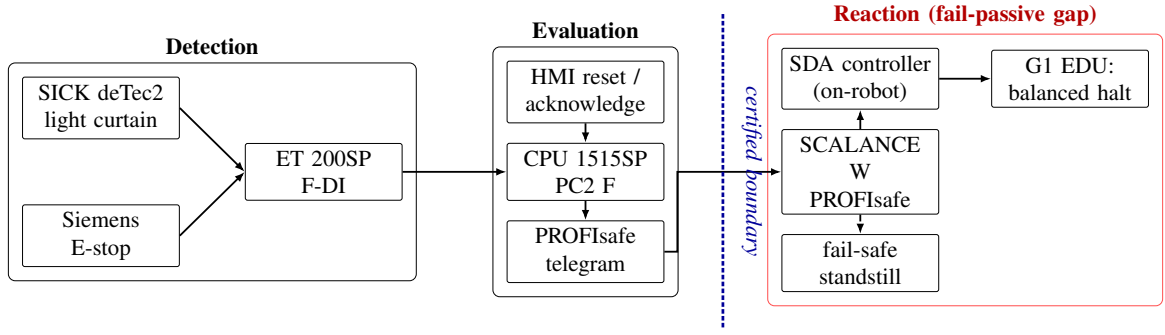
\begin{figure*}[t]
\centering
\resizebox{0.86\textwidth}{!}{%
\begin{tikzpicture}[
    font=\normalsize,
    box/.style={draw, rounded corners=1pt, align=center, inner sep=3pt,
                minimum height=8mm, text width=22mm, fill=white},
    grp/.style={draw, rounded corners=3pt, inner sep=6pt},
    flow/.style={-{Latex[length=1.6mm]}, thick},
    wd/.style={-{Latex[length=1.6mm]}, thick, dashed},
    node distance=3mm and 4mm
]
\node[box] (lc) {SICK deTec2\\light curtain};
\node[box, below=10mm of lc] (es) {Siemens\\E-stop};
\node[box, right=10mm of lc.east |- {$(lc)!0.5!(es)$}, anchor=west] (fdi) {ET 200SP\\F-DI};
\node[box, right=16mm of fdi] (cpu) {CPU 1515SP\\PC2 F};
\node[box, above=of cpu] (rst) {HMI reset /\\acknowledge};
\node[box, below=of cpu] (ps) {PROFIsafe\\telegram};
\node[box, right=18mm of cpu] (wlan) {SCALANCE W\\PROFIsafe};
\node[box, above=of wlan] (rx) {SDA controller\\(on-robot)};
\node[box, below=of wlan] (fs) {fail-safe\\standstill};
\node[box, right=8mm of rx] (halt) {G1 EDU:\\balanced halt};
\draw[flow] (lc.east) -- (fdi.west);
\draw[flow] (es.east) -- (fdi.west);
\draw[flow] (fdi.east) -- (cpu.west);
\draw[flow] (rst.south) -- (cpu.north);
\draw[flow] (cpu.south) -- (ps.north);
\draw[flow] (ps.east) -- ++(2mm,0) |- (wlan.west);
\draw[flow] (wlan.north) -- (rx.south);
\draw[flow] (rx.east) -- (halt.west);
\draw[wd] (wlan.south) -- (fs.north);
\begin{scope}[on background layer]
\node[grp, fit=(lc)(es)(fdi), label=above:{\bfseries Detection}] (det) {};
\node[grp, fit=(rst)(cpu)(ps), label=above:{\bfseries Evaluation}] (ev) {};
\node[grp, draw=red!70, fit=(rx)(wlan)(fs)(halt),
      label=above:{\bfseries\color{red!70!black} Reaction (fail-passive gap)}] (re) {};
\end{scope}
\draw[very thick, blue!60!black, dash pattern=on 2pt off 1pt]
    ($(ev.north east)!0.5!(re.north west) + (0,4mm)$) --
    ($(ev.south east)!0.5!(re.south west) + (0,-4mm)$)
    node[midway, right, rotate=-90, anchor=south, text=blue!60!black,
         yshift=1.5mm]{\itshape certified boundary};
\end{tikzpicture}%
}
\caption{D--E--R safety supervision architecture. The certified external chain
(Detection, Evaluation) acts as the instrument; the Reaction chain (red) is the
fail-passive gap. Robot-side reception runs on the SDA controller, relocating the
gap to the SDA-to-balancing-policy interface. Dashed arrow: watchdog-triggered
fail-safe standstill on communication loss.}
\label{fig:arch}
\end{figure*}

The safety function is organized as three subsystems (Fig.~\ref{fig:arch}):
Detection (light curtain, E-stop, F-DI), Evaluation (fail-safe CPU, decision
logic, reset handling, PROFIsafe telegram supervision), and Reaction (wireless
PROFIsafe over SCALANCE~W, robot-side reception on the SDA controller, transition to
a balanced standing halt, and fail-safe standstill on communication loss). A
distinguishing feature is that the robot-side PROFIsafe endpoint is a
software-defined automation (SDA) controller (a soft PLC) running in a container on
the G1's onboard Linux compute:
the stop telegram is received and handled by a standardized IEC~61131-3 runtime
located \emph{on the robot}, which then commands the native locomotion policy to
enter the balanced halt. Because this runtime executes on the G1's standard
(non-safety-rated) onboard compute, it standardizes the reception \emph{interface}
but is not a certified PROFIsafe F-host (Sec.~\ref{sec:eval}). The system defines
\emph{one safety zone} covering the accessible open side; any zone violation or
E-stop actuation produces a stop demand. Multi-zone speed-and-separation monitoring
is deferred to future work.

\section{Safety Function Design}

\subsection{Safety Requirement Specification}
\textbf{SF-1:} \emph{When the protected open side is intruded, or when the
emergency stop is actuated, the robot shall be commanded to a safe stop and shall
reach a stable balanced standstill; the system shall not permit motion to resume
until a manual reset is acknowledged.} Trigger conditions: (1) a human or the
robot interrupts the light curtain; (2) the emergency-stop device is pressed.
Manual reset/acknowledgement (via HMI) is required after any stop event. On
communication loss the robot stands still (fail-safe standstill via PROFIsafe
watchdog timeout). The safe state after a stop is standing statically with
balance (not a de-energized collapse). The emergency stop is a supplementary
safety function per EN~ISO~13850; primary risk reduction is the light-curtain
protective stop plus the workspace layout.

\subsection{Reset, Restart, and Auxiliary LiDAR}
After any stop the safety logic latches; resumption requires the operator to clear
the intrusion / release the E-stop and acknowledge via the HMI, satisfying the
``no automatic restart'' principle of EN~ISO~13850 and EN~60204-1. Separately, we
implement an onboard LiDAR-based local obstacle stop that runs on the robot as an
\emph{auxiliary} collision-avoidance behavior. We deliberately keep it outside the
certified argument: it is not a certified safety function---its PL, DC, and
independence are not established---and it is excluded from the SIL/PL argument,
complementing rather than substituting for the external protective stop.

\subsection{Stop Category Discussion}
\label{sec:stopcat}
The contrast with the reference application example (Sec.~\ref{sec:ref}) is
instructive. There, the emergency stop switches off actuators via contactors
according to Stop Category~0 (immediate removal of power). The humanoid cell
deliberately does not do this: removing power from a walking biped would cause an
uncontrolled fall. Instead the system commands a controlled standing halt while
maintaining the control authority needed to remain balanced. The implemented
response is therefore \emph{not} Stop Category~0; it more plausibly resembles a
controlled stop with power maintained to achieve and hold the stop, tentatively
resembling Stop Category~1 (EN~60204-1). We adopt cautious wording: a definitive
classification requires system-level verification of the robot's internal stop
dynamics and of whether/when power is ultimately removed, which is outside the
certified boundary of the external chain.

\section{Standards-Based Safety Evaluation Methodology}
\label{sec:eval}

\subsection{Applicable Standards}
ISO~13849-1 (EN~ISO~13849-1:2015) is the primary quantitative framework; ISO~13849-2 provides
validation; IEC~62061 provides the parallel SILCL assessment. EN~ISO~13850
governs the emergency-stop function; EN~60204-1 governs stop categories.
ISO~13855 governs protective-equipment placement (Sec.~\ref{sec:sep}); IEC~61496
type-rates the light curtain. ISO~10218:2025 and ISO/TS~15066 (SSM/PFL) are used
for contrast, since they assume a fail-passive safe state. These standards apply
directly to the Detection and Evaluation subsystems; they are only partially
adaptable to Reaction, whose legged stop dynamics have no governing machinery
standard.

\subsection{Quantitative Method}
The core parameters ISO~13849-1 uses to quantify a safety function are category,
MTTF$_\mathrm{D}$, DC, CCF, and PFHD $\rightarrow$ PL; the arrow means the first
five \emph{combine to determine} the last one (the PL). For each subsystem we
follow the ISO~13849-1 procedure:
determine the
architecture \emph{category} (structural/redundancy class B,\,1--4; reference:
Category~4; external chain targets Category~3/4); establish
\emph{MTTF$_\mathrm{D}$} (mean time to dangerous failure) from component data
(B10, dangerous-failure fraction, mission time); establish \emph{DC} (diagnostic
coverage---the fraction of dangerous failures detected; reference $\geq99\%$ via
F-DI cross-comparison); verify \emph{CCF} (common-cause-failure resistance: the
redundant subsystems pass the ISO~13849-1 qualitative gate at $\geq65$ points per
Table~F.1 and take a conservative 10\% common-cause factor, $\beta=0.1$, in the
IEC~62061 quantitative cross-check); compute
\emph{PFHD} (dangerous failures per hour) and map to the \emph{PL} (Performance
Level a--e); and cross-check with IEC~62061 SILCL. These inputs are named because
the reference Reaction subsystem is two electromechanical, cycle-rated contactors:
for such parts MTTF$_\mathrm{D}$ is built up from the B10 cycle life, the
dangerous-failure fraction, and the mission time, whereas purely electronic parts
are given a direct failure rate.

\subsection{Methodological Reference Values}
\label{sec:ref}
To illustrate the method---not as recalculated values for the humanoid reaction
chain---we cite the Siemens application example \emph{``Emergency Stop up to PL~e
/ SIL~3 with a Fail-Safe S7-1500 Controller''}~\cite{siemens21064024}. It uses the same D--E--R decomposition. Table~\ref{tab:iso} gives the
results (both frameworks).

\begin{table}[htbp]
\caption{Reference subsystem PFHD under ISO 13849-1 (PL) and IEC 62061 (SILCL)}
\label{tab:iso}
\begin{center}
\footnotesize
\renewcommand{\arraystretch}{1.1}
\setlength{\tabcolsep}{3pt}
\resizebox{\columnwidth}{!}{%
\begin{tabular}{|p{2.7cm}|c|c|c|c|}
\hline
\textbf{Subsystem (reference)} & \textbf{PFHD} & \textbf{PL} & \textbf{PFHD} & \textbf{SILCL}\\
 & \multicolumn{2}{c|}{ISO 13849-1} & \multicolumn{2}{c|}{IEC 62061}\\
\hline
Detection & $9.06\!\times\!10^{-10}$ & e & $1.19\!\times\!10^{-10}$ & 3\\
\hline
Evaluation & $5.00\!\times\!10^{-9}$ & e & $5.00\!\times\!10^{-9}$ & 3\\
\hline
Reaction (2 contactors, Cat.~0) & $1.45\!\times\!10^{-9}$ & e & $7.30\!\times\!10^{-9}$ & 3\\
\hline
\textbf{Total} & $\mathbf{7.35\!\times\!10^{-9}}$ & \textbf{e} & $\mathbf{1.24\!\times\!10^{-8}}$ & \textbf{3}\\
\hline
\end{tabular}%
}
\end{center}
\end{table}

Key assumptions: Detection uses an E-stop device (Category~4, DC~$\geq99\%$ via
F-DI cross-comparison, 1oo2 equivalent, CCF~$\geq65$ points / $\beta=0.1$).
Evaluation sums F-CPU incl.\ PROFIsafe ($2.00\times10^{-9}$) + F-DI
($1.00\times10^{-9}$) + F-DQ ($2.00\times10^{-9}$). Reaction uses two contactors
(Category~4, DC~$\geq99\%$ via redundant switch-off and dynamic monitoring). These
figures show a detection+evaluation chain of this class is capable of PL~e /
SIL~3; they are the reference's component-specific values, \emph{not} claims about
the humanoid reaction chain. A valid rating for the actual G1 cell would require
recomputing each subsystem's PFHD from the failure data of the specific safety
devices installed here.

\subsection{The Reaction-Subsystem Substitution}
The reference certifies its Reaction subsystem by hardware power removal: two
monitored contactors achieve PFHD~$\approx1.45\times10^{-9}$ (PL~e) in a Stop
Category~0. In the humanoid cell there are no contactors; the identical logical
position in the D--E--R chain is occupied by the wireless PROFIsafe path plus the
robot's control policy, which brings the platform to an actively balanced
standstill rather than removing power. Thus the exact subsystem that is
straightforwardly certifiable in the reference becomes uncertified and
unstandardized in the humanoid---certified contactor power-removal $\rightarrow$
uncertified legged controlled-halt---the precise technical locus of the
fail-passive gap.

Deploying the SDA controller on the robot standardizes, but does not certify, the
front of this chain (telegram reception, watchdog supervision, decision handling)
via a well-defined industrial IEC~61131-3 runtime. Crucially, a genuine PROFIsafe
F-host (with a SIL/PL claim) would require \emph{safety-rated} hardware; the G1's
onboard compute is standard commercial off-the-shelf (COTS), so the SDA controller is \emph{not} a certified
safety runtime and its reception element carries no PFHD/DC/CCF claim. This does not close the
gap---it makes its location precise: the residual, uncertified element is the
\emph{SDA-to-balancing-policy interface} on non-safety-rated compute
(Sec.~\ref{sec:limits}).

\subsection{Certification Claim Scope}
We claim the architecture \emph{targets} PL~e / SIL~3-capable components and
methods, and that the PLC-side and sensor-side chain \emph{can be evaluated} using
established methods. We do \emph{not} claim certified PL~e / SIL~3 for the complete
humanoid system, nor a validated PFHD for the robot-side reaction chain. The
end-to-end deployment is a feasibility-oriented certification study.

\subsection{Protective Separation Distance (ISO 13855)}
\label{sec:sep}
The placement of the light curtain on the open side is governed by ISO~13855:
\begin{equation}
S = K \cdot T + C
\end{equation}
where $S$ is the minimum separation from the detection plane to the hazard, $K$
the approach-speed constant that the standard fixes ($1600\,\mathrm{mm/s}$ for a walking body,
$2000\,\mathrm{mm/s}$ for hand/arm reach), $T=t_\mathrm{response}$
the total system stopping time (Sec.~\ref{sec:timing}), and $C=8(d-14)\geq0$ the
intrusion-depth term for detection capability $d$ (mm). Two humanoid-specific
caveats apply: (i)~$T$ is dominated by the uncertified reaction chain
($t_\mathrm{stop}$ depends on the balancing policy and gait phase), so $S$ can only
be \emph{bounded} using worst-case measured $t_\mathrm{stop}$; and (ii)~the hazard
can move toward the curtain (a walking humanoid may approach the boundary), so the
effective closing speed is the sum of human and robot approach, bounded here by
constraining the waypoints and commanded speed. We therefore report $S$ as a
worst-case bounded value justifying the mounting offset, not a certified
separation.

\section{Experimental Setup}

\begin{figure*}[t]
\centering
\begin{minipage}[t]{0.24\textwidth}\centering
  \includegraphics[width=\linewidth]{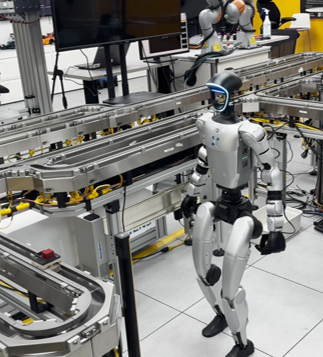}\\[1pt]{\small (a) Approach}
\end{minipage}\hfill
\begin{minipage}[t]{0.24\textwidth}\centering
  \includegraphics[width=\linewidth]{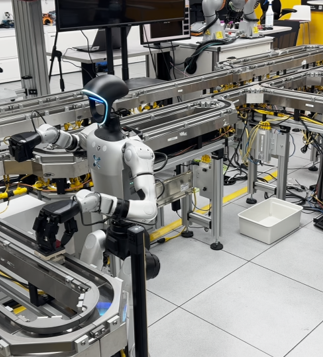}\\[1pt]{\small (b) Pick-up}
\end{minipage}\hfill
\begin{minipage}[t]{0.24\textwidth}\centering
  \includegraphics[width=\linewidth]{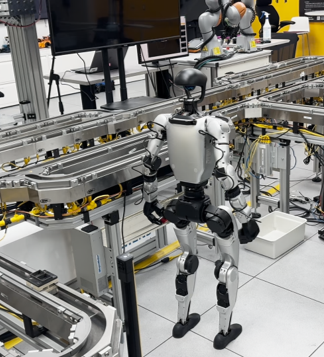}\\[1pt]{\small (c) Transport}
\end{minipage}\hfill
\begin{minipage}[t]{0.24\textwidth}\centering
  \includegraphics[width=\linewidth]{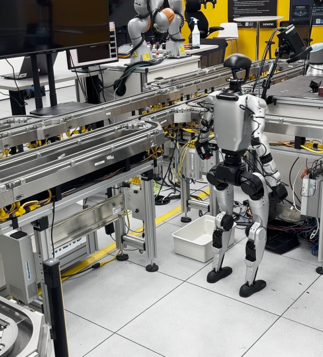}\\[1pt]{\small (d) Drop}
\end{minipage}
\caption{Demonstrator cube-handling task executed by the humanoid:
(a)~approach, (b)~pick-up, (c)~transport, and (d)~drop. Intrusion and E-stop
scenarios (S1--S5) are injected at each phase.}
\label{fig:taskflow}
\end{figure*}

\subsection{Objectives and Value Classes}
The program (i) characterizes external-chain timing, (ii) empirically bounds the
wireless PROFIsafe reaction path, and (iii) exposes what is not yet known about
the robot-side reaction chain. Every reported quantity is labeled by provenance:
\textbf{[S]} specified, \textbf{[C]} configured, or \textbf{[M]} measured.

\subsection{Test Scenarios and Trial Plan}
Each scenario injects a disturbance into the cube-handling task flow
(Fig.~\ref{fig:taskflow}) over 20--30 trials ($\geq10$ if resource-limited); we
report best/average/worst-case statistics. Scenarios: S1 intrusion during
walking; S2
intrusion during pick; S3 intrusion during drop; S4 E-stop during walking; S5
E-stop during manipulation; S6 communication loss during motion (fail-safe
standstill); S7 repeated reset/restart; S8 nuisance-trigger observation.

\subsection{Instrumentation}
TIA Portal trace (PLC I/O timing); PLC diagnostics/timestamps (scan time,
F-telegram events); robot controller logs (reception, policy-state transitions);
synchronized high-frame-rate video (robot stopping time/distance, time-to-stable
posture); SCALANCE~W network timing logs (latency, packet loss). A shared time
base aligns each electrical event (detection, F-telegram, reception) with the
optically measured mechanical stop, decomposing $t_\mathrm{response}$
(Sec.~\ref{sec:timing}) into per-stage terms.

\subsection{System-Level Timing/Latency Budget}
\label{sec:timing}
The end-to-end response time is
\begin{equation}
t_\mathrm{response} = t_\mathrm{detect} + t_\mathrm{FDI} + t_\mathrm{scan} +
t_\mathrm{PROFIsafe} + t_\mathrm{rx} + t_\mathrm{stop}.
\end{equation}
In order, these are light-curtain detection, the F-DI input delay, the
safety-PLC scan, PROFIsafe transmission (watchdog plus WLAN latency),
robot-side reception, and the mechanical stop.
Each term is labeled by provenance: \textbf{[S]} specified (data sheet),
\textbf{[C]} configured (TIA Portal), or \textbf{[M]} measured.
Table~\ref{tab:timing} reports the budget terms for the single-cell
demonstrator.

\begin{table}[htbp]
\caption{Timing/latency budget for the single-cell demonstrator. \textbf{Prov.}\
marks each term's source ([S] specified, [C] configured, [M] measured).}
\label{tab:timing}
\begin{center}
\footnotesize
\renewcommand{\arraystretch}{1.1}
\begin{tabular}{|c|c|p{2.7cm}|c|}
\hline
\textbf{Term} & \textbf{Prov.} & \textbf{Source} & \textbf{Value}\\
\hline
$t_\mathrm{detect}$ & [S] & deTec2 Core response time & $11$~ms\\
\hline
$t_\mathrm{FDI}$ & [C] & F-DI delay + filter & $4$--$10$~ms\\
\hline
$t_\mathrm{scan}$ & [M] & PLC F-runtime cycle & $15$--$40$~ms\\
\hline
$t_\mathrm{PROFIsafe}$ & [C]/[M] & WLAN latency; $F\_WD\_Time$ & $30$--$39$; $\le192$~ms\\
\hline
$t_\mathrm{rx}$ & [M] & robot controller log & $5$--$20$~ms\\
\hline
$t_\mathrm{stop}$ & [M] & synchronized video & $0.3$--$1.0$~s\\
\hline
\end{tabular}
\end{center}
\end{table}

The certifiable worst-case bound sums worst-case terms; for the external chain the
dominant configured bound is the PROFIsafe watchdog $F\_WD\_Time$, which
upper-bounds $t_\mathrm{PROFIsafe}$ and triggers the fail-safe standstill on
communication loss (S6). Because the demonstrator confines the robot to a single
access-point cell (the $3\times1.5$~m workspace, no roaming), the dominant roaming
contribution to wireless latency is removed, admitting a tighter $F\_WD\_Time$---hence
a smaller separation; multi-access-point roaming is left to future work.
Summing the worst-case terms gives
$t_\mathrm{response}^\mathrm{wc}\approx1.1$~s (dominated by $t_\mathrm{stop}$),
and the worst-case advance toward the hazard used in the ISO~13855 calculation is
\begin{equation}
d_\mathrm{stop}^\mathrm{wc}(v) = v\, t_\mathrm{response}^\mathrm{wc} +
d_\mathrm{decel}(v).
\end{equation}
Only $t_\mathrm{detect}\ldots t_\mathrm{PROFIsafe}$ lie within the
conventionally-evaluable external chain; $t_\mathrm{rx}$ and $t_\mathrm{stop}$ lie
inside the uncertified reaction chain and can be measured but not certified. This
partition mirrors the fail-passive gap.

\section{Results}

\emph{Status: the timing budget (Table~\ref{tab:timing}) is populated from
specified, configured, and measured values, and the communication-loss fail-safe
standstill is demonstrated; the remaining reliability and stopping-distance results
are future work. No unsupported numbers are asserted.} Timing metrics
($t_\mathrm{detect}$, E-stop response, $t_\mathrm{scan}$,
$t_\mathrm{PROFIsafe}$, $t_\mathrm{stop}$, time-to-stable-posture) follow
Table~\ref{tab:timing}. Communication loss (S6), induced by four independent
methods---access-point power-off, antenna disconnect, wired-uplink disconnect, and
F-CPU stop (three trials each)---drove the robot to a stable balanced standstill
(no topple) in $0.5$--$1.3$~s in every trial, matching the
$F\_WD\_Time + t_\mathrm{stop}$ budget. Nuisance-trip rate (S8), availability (S7),
the packet-loss watchdog margin, and a full stopping-distance curve
$d_\mathrm{stop}(v)$ are the subject of a dedicated follow-up study.
Feasibility of PFHD/DC/CCF evaluation on the external chain is supported by
Sec.~\ref{sec:ref}; crucially, the reference closes its overall figure only because
its Reaction subsystem is monitored contactors ($\approx1.45\times10^{-9}$), so
with no such element the humanoid chain has no end-to-end PFHD yet. A formal
recalculation for this cell is future work.

\section{Discussion}

The external supervision architecture is buildable entirely from certified
components (no custom safety hardware), and the D--E decomposition maps cleanly
onto ISO~13849-1 / IEC~62061 procedures; wireless PROFIsafe is architecturally
adequate for safe stop signaling, with the watchdog timeout setting the fail-safe
standstill latency (S6, demonstrated), subject to packet-loss validation. The
reaction chain is where certification currently stops: the active safe state
depends on the robot's proprietary control policy, whose PFHD, DC, and fault
behavior are not established, and (Sec.~\ref{sec:eval}) the on-robot SDA reception
endpoint is itself uncertified on non-safety-rated compute. Overclaiming PL~e /
SIL~3 for the full humanoid would be unsupportable.

\subsection{Humanoid-Specific Safety Analysis of the Active Safe State}

\textbf{Fall-as-hazard trade-off.} For a fail-passive machine ``stop'' is
unconditionally safe. For the humanoid it is non-monotonic: an overly aggressive
halt can inject a disturbance exceeding the balance controller's recovery margin
and induce a fall, converting a protective action into a new hazard. The safe stop
is thus a \emph{constrained} stop---minimize stopping time/distance while remaining
within the capturable region---with no analog in ISO~13849 / EN~60204-1.

\textbf{Mid-step (single-support) demands.} The outcome depends on gait phase. In
double-support the support polygon is large and deceleration is quick; if the
demand arrives in single-support (swing), immediate freezing is generally
infeasible and the controller must place the current step (reach a capture point)
before holding a static posture. This imposes a phase-dependent lower bound on
$t_\mathrm{stop}$, so the ISO~13855 distance must be sized using the worst-case
(single-support) $t_\mathrm{stop}$.

\textbf{Residual risk of balancing-policy failure.} The active safe state must be
\emph{held} by the control policy, so failure of that policy (software fault,
sensor dropout, actuator saturation, or a disturbance beyond the recovery margin)
during or after the stop is a distinct hazard; a fail-passive safe state, by
contrast, persists without energy or computation. This residual risk has no PFHD
characterization and no governing standard; interim mitigation is qualitative
(bound the demand rate, restrict walking speed to the recovery envelope, treat
balancing-policy faults as fall hazards requiring physical mitigation). A
complementary runtime approach learns a safe-stoppability monitor that predicts
whether the current state can still reach a balanced minimum-risk condition and
triggers the fallback before that becomes infeasible~\cite{sun2026learning};
this can shrink the residual risk but, being data-driven, does not by itself
furnish a certifiable PFHD/DC argument. Finally,
the ISO~13855 distance $S=K\,T+C$ depends on $T$, dominated by the
phase-dependent, uncertified $t_\mathrm{stop}$; even a standard placement
calculation thus inherits the fail-passive gap.

\section{Limitations and Certification Gaps}
\label{sec:limits}
\begin{itemize}
\setlength{\itemsep}{0pt}\setlength{\parskip}{0pt}
\item \textbf{Robot-side reaction chain (primary gap).} In the reference, Reaction
is two monitored contactors achieving PFHD~$\approx1.45\times10^{-9}$ (PL~e /
SILCL~3) by Stop Category~0 power removal. In the humanoid, that certifiable
element is replaced by wireless PROFIsafe reception (on an SDA controller) plus the
robot's proprietary balancing policy; the SDA reception \emph{interface} is
standardized, but motion-policy interruption and transition to a balanced
standstill are not certified, no PFHD/DC/CCF is established, and no
humanoid-specific standard governs them.
\item \textbf{No safety-rated onboard compute.} A genuine PROFIsafe F-host on the
robot would require safety-rated hardware; the G1's onboard compute is standard
COTS, so the SDA endpoint is \emph{not} a certified safety runtime and carries no
SIL/PL claim. A certified robot-side endpoint (safety-rated compute plus a formal
PFHD/DC/CCF evaluation) and the integrity of the SDA-to-balancing-policy hand-off
remain open.
\item \textbf{Safe-state \& stop category.} ``Standing statically with balance''
is an actively maintained state, not fail-passive; the controlled halt tentatively
resembles Stop~Cat.~1 but is unverified (Sec.~\ref{sec:stopcat}).
\item \textbf{Reference figures} are methodological, not recalculated for this
cell. Wireless latency is measured and the comms-loss fail-safe standstill is
demonstrated (four interruption methods); only the packet-loss watchdog margin
(S6) remains for the follow-up study.
\item \textbf{Scope.} Single zone, single scenario (may not generalize);
manipulation-phase interruption may leave the workpiece indeterminate.
\end{itemize}

\section{Conclusion}

The safe state of a legged robot is an actively-controlled state, which violates
the fail-passive assumption underlying classical machinery safety (ISO~13849-1 /
EN~60204-1). This fail-passive gap is why industrial humanoid deployment cannot
yet reach end-to-end certification, and characterizing it precisely was the goal
of this paper. We built a conventional, certified external safety supervision
chain around a Unitree G1 EDU pick-and-place cell---Detection, Evaluation, and
Reaction over wireless PROFIsafe---and used it as an instrument. Because the
external chain is closed and quantifiable with established methods (as evidenced by
the Siemens fail-safe S7-1500 reference reaching PL~e / SIL~3, with its Reaction
subsystem certifiable precisely because it is contactor-based Stop Category~0), the
residual uncertifiable element is pinpointed to one place: the robot-side reaction
chain, where an active safe state replaces contactor power removal. Our positive
contributions (C1--C5) are concrete, while we explicitly do not claim certified
PL~e / SIL~3 for the complete humanoid. The appropriate framing for present
industrial-humanoid safety work is therefore certification-oriented feasibility and
gap analysis---a reproducible blueprint paired with a precise research agenda
toward closing the fail-passive gap.

Future work will (i) perform a hardware-specific PFHD/DC/CCF calculation for the
detection and evaluation subsystems, and pursue a certified robot-side endpoint
(SDA safety controller on safety-rated onboard compute); (ii) establish the
reaction chain's fault behavior and diagnostic coverage (timing now measured);
(iii) develop a defensible stop-behavior model to classify the controlled halt;
(iv) extend the single-zone architecture to speed-and-separation monitoring; and
(v) characterize reliability (nuisance-trip rate, packet loss, availability).


\bibliographystyle{IEEEtran}
{\footnotesize
\bibliography{references}
}

\end{document}